\documentclass[conference]{IEEEtran}
\IEEEoverridecommandlockouts
\usepackage{cite}
\usepackage{amsmath,amssymb,amsfonts}
\usepackage{algorithmic}
\usepackage{graphicx}
\usepackage{textcomp}
\usepackage{xcolor}

\usepackage{colortbl}
\usepackage{boldline}
\usepackage[flushleft]{threeparttable}
\usepackage{multirow}
\usepackage{array}
\usepackage{algorithm}

\usepackage{times}
\usepackage{soul}
\usepackage{url}
\usepackage[utf8]{inputenc}
\usepackage[small]{caption}
\usepackage{graphicx}
\usepackage{amsmath}
\usepackage{amsthm}
\usepackage{booktabs}
\usepackage{colortbl}
\usepackage{subcaption}
\usepackage{boldline}

\newcommand{\norm}[1]{\left\lVert#1\right\rVert}
\newcommand{\inv}{^{\text{-}1}}

\usepackage{xspace}

\def\BibTeX{{\rm B\kern-.05em{\sc i\kern-.025em b}\kern-.08em
    T\kern-.1667em\lower.7ex\hbox{E}\kern-.125emX}}
\begin{document}

\title{MARLINE: Multi-Source Mapping Transfer Learning for Non-Stationary Environments\\
\thanks{Authors in order of contribution.}
\thanks{L. Minku was supported by EPSRC Grant No. EP/R006660/2 and H. Zhou was supported by EU Horizon 2020 DOMINOES Project (grant number: 771066).}
}

\author{
\IEEEauthorblockN{Honghui Du}
\IEEEauthorblockA{\textit{School of Informatics} \\
\textit{University of Leicester}\\
Leicester, United Kingdom \\
hd168@leicester.ac.uk}
\and
\IEEEauthorblockN{Leandro L. Minku}
\IEEEauthorblockA{\textit{School of Computer Science} \\
\textit{University of Birmingham}\\
Birmingham, United Kingdom \\
L.L.Minku@cs.bham.ac.uk}
\and
\IEEEauthorblockN{Huiyu Zhou}
\IEEEauthorblockA{\textit{School of Informatics} \\
\textit{University of Leicester}\\
Leicester, United Kingdom \\
hz143@leicester.ac.uk}
}

\maketitle

\begin{abstract}
Concept drift is a major problem in online learning due to its impact on the predictive performance of data stream mining systems. Recent studies have started exploring data streams from different sources as a strategy to tackle concept drift in a given target domain. These approaches make the assumption that at least one of the source models represents a concept similar to the target concept, which may not hold in many real-world scenarios. In this paper, we propose a novel approach called Multi-source mApping with tRansfer LearnIng for Non-stationary Environments (MARLINE). MARLINE can benefit from knowledge from multiple data sources in non-stationary environments even when source and target concepts do not match. This is achieved by projecting the target concept to the space of each source concept, enabling multiple source sub-classifiers to contribute towards the prediction of the target concept as part of an ensemble. Experiments on several synthetic and real-world datasets show that MARLINE was more accurate than several state-of-the-art data stream learning approaches. 
\end{abstract}

\begin{IEEEkeywords}
concept drifts, non-stationary environment, multi-sources, transfer learning.
\end{IEEEkeywords}

\section{Introduction}
The need for efficient streaming data analytics has rapidly grown in recent years \cite{lu2018learning}.
A data stream can be defined as a sequence of observations that continuously arrive over time, occurring in many applications, such as credit card approval, fraud detection, and software defect prediction \cite{du2019multi}. A key challenge in data stream learning is that the joint probability distribution of an application may change over time, i.e., there may be concept drift \cite{pocock2010online}. Learning from data streams that may suffer from concept drifts is frequently referred to as learning in non-stationary environments \cite{ditzler2015learning,minku2019transfer}, whereas a given joint probability distribution can be treated as a concept \cite{du2019multi,gama2014survey}. Data stream learning algorithms must be able to adapt and swiftly react to concept drifts to avoid poor predictions \cite{lu2018learning}.

Using information learned from different data sources is a feasible way to speed up the learning of a new target concept and improve the accuracy of the predictions. This can be considered as transfer learning \cite{pan2010survey}. However, transfer learning has been usually used off-line, requiring the entire training set to be available before training commences. While a few recent studies applied transfer learning in non-stationary data streaming environments \cite{minku2019transfer}, most of the approaches presume a similarity existing between the source and target concepts \cite{du2019multi, dong2019multistream, tao2019comc}. This assumption often fails to hold in practice. For example, the concept underlying the prediction for bike rental demands in Washington D.C.~and London are different due to different weather patterns and consumer behaviours in these two cities. However, data streams of these two locations are available \cite{LondonBike:2019,fanaee2014event} and could potentially be used to improve the predictive performance of data stream learning approaches. 
Another example is software effort estimation, where data streams describing software projects developed by different companies may be used to improve software effort estimation in a given company, despite having different underlying distributions \cite{minku2014make}.  

Therefore, this paper aims to answer the following research questions:
\textit{Can multi-source transfer learning help us to improve the predictive performance in non-stationary environments where source and target data streams do not share the same concept? If so, how?}

To answer these questions, we hereby propose a novel approach, namely Multi-source mApping tRansfer LearnIng for Non-stationary Environments (MARLINE). MARLINE is the first approach designed to benefit from multiple source data streams even when sources and target could have considerably different concepts. 
It achieves that by projecting the target concept to the space of each source concept through a novel mapping mechanism, enabling multiple source sub-classifiers to contribute towards the prediction of the target concept as part of an ensemble. 
Our experiments show that MARLINE can improve the predictive performance of the existing approaches over time and quickly obtain good performance at the early learning stage or after the concept drift occurs, even though there are only a few training examples available. 


\section{Related Work}
\label{sec:RelatedWork}


Several approaches have been proposed for data stream learning in non-stationary environments \cite{gama2014survey,ditzler2015learning}. Among these, approaches able to learn example-by-example (online) rather than chunk-by-chunk \cite{krawczyk2017ensemble} are particularly relevant to our paper. They have the potential to adapt to concept drifts faster than  chunk-based approaches. Such approaches can be further divided into active and passive approaches \cite{ditzler2015learning,krawczyk2017ensemble}.
Active approaches trigger the adaptation mechanisms when concept drift detection methods alert \cite{ditzler2015learning}, \cite{gomes2017adaptive}. Examples of drift detection methods include the Drift Detection Method (DDM) \cite{gama2004learning} and Drift Detection Methods based on the Hoeffding’s inequality (HDDM) (HDDM$_A$ and HDDM$_W$) \cite{frias2014online}. Passive approaches continuously adapt to concept drift without relying on explicit concept drift detection \cite{ditzler2015learning,krawczyk2017ensemble}.  A popular passive approach is Dynamic Weighted Majority (DWM) \cite{kolter2007dynamic}. In spite of their learning capacity, none of these approaches uses transfer learning 
or operates in multi-source scenarios. 

Very few approaches have used transfer learning in non-stationary environments \cite{minku2019transfer}. Online inductive parameter transfer learning approaches include Dynamic Cross-company Mapped Model Learning (Dycom) \cite{minku2014make}, Diversity for Dealing with Drifts (DDD) \cite{minku2011ddd} and Online Window Adjustment Algorithm (OWA) \cite{zhao2014online}. Dycom treats source data offline, whereas DDD and OWA do not use source data, transferring knowledge only from the immediate previous target concept to the current target concept.
A new chunk-based inductive parameter transfer approach called Diversity and Transfer-based Ensemble Learning (DTEL) was proposed in \cite{sun2018concept}. Similar to DDD, DTEL does not consider different sources but uses historical target concepts. In addition, this is a chunk-based approach. 
Recently, two transductive transfer learning approaches called MultiStream Classification using Relative Density Ratio (MSCRDR) \cite{dong2019multistream} and Cross-domain Multistream Classification (COMC) \cite{tao2019comc} have been proposed to handle multiple sources, with the assumption that the target stream is unlabelled and the source is labelled. However, these two algorithms require both the source and the target to share the same task, even after a concept drift happens. Multi-source transfer learning for non-stationary environments (Melanie) \cite{du2019multi} is another online transfer learning method that can learn from multiple sources. However, Melanie only benefits from source concepts that are similar to the target.

Overall, there is no existing approach to perform transfer learning using multiple non-stationary data sources that may have different concepts from those of the target.

\section{Problem Statement}
\label{sec:ProblemStatement}

Let $\{\textbf{x}_i, y_i\}$ denote an example received at a given point in time at a data stream $i$ with domain $\mathcal{D}_i = \{ \mathcal{X}_i, p_i(\textbf{x}) \}$ and task $\mathcal{T}_i = \{ \mathcal{Y}_i, p_i(y|\textbf{x}) \}$, where $i \in \{S_1, S_2, \cdots, S_n, T\}$, $S_n$ is the $n$th source data stream, $T$ is the target data stream, $x_i \in \mathcal{X}_i$, $\mathcal{X}_i$ is a $d$-dimensional feature space, $y_i \in \{-1, +1\}$ is the class label, $p_i(\textbf{x})$ is the marginal probability distribution and $p_i(y | \textbf{x})$ is the posterior probability distribution. 

All sources and target streams may suffer concept drift. 
We will enumerate the concepts $p_i^j$ seen in a data stream $i$ using a sequential identifier $j$. Whenever a concept drift occurs, we increment $j$. We use $J_i$ to denote the number of concepts observed so far in data stream $i$. Note that a given training example, domain and task are all associated with the given concept. Therefore, they are actually all indexed by $j$ as shown in $\{\textbf{x}_i^j, y_i^j\}$, $\mathcal{D}_i^j$, and $\mathcal{T}_i^j$, but we will leave this index implicit. 

At the beginning of the data stream or after a concept drift, due to the lack of the target data representing the new concept, the performance of the predictive models is usually poor. Therefore, the aim of  multi-source transfer is to improve predictive performance in non-stationary environments and speed up learning especially in the beginning of the data stream or after the occurrence of concept drifts by using the data from multiple sources. We will investigate inductive transfer learning (i.e. $\mathcal{T}_{S_i} \neq \mathcal{T}_T$ while $\mathcal{D}_{S_i} \neq \mathcal{D}_T$ or $\mathcal{D}_{S_i} = \mathcal{D}_T$), as concept drifts may cause changes in $\mathcal{T}_T$ and $\mathcal{T}_{S_i}$ over time.

\begin{table}[t]\centering
\caption{Key Notation and Description.}
\vspace{-0.2cm}
\scalebox{1}{
\begin{tabular}{|c| p{6.5cm}|}
\hline
$S_n$ &The data stream from source $n$ \\
$T$ &The target data stream \\
$\mathcal{M}$ & Set of sources and target data streams seen so far\\
$\textbf{x}_i$ & Feature vector from data stream $i$\\
$y_i$ & The corresponding label of $\textbf{x}_i$\\
$d$ & Dimensionality of the feature space\\
$J_i$ & Number of concepts from data stream $i$ observed so far \\
$p^j_i$ & The $j$th concept of data stream $i$, $0< j \leq J_i$\\
$\textbf{x$'$}_i^{j}$ & The projection of $\textbf{x}_T$ on the $j$th concept of stream $i$\\
$H_i^{j}$ & The base learning ensemble classifier trained by the $j$th $concept$ of data stream $i$\\
$H_i$ & Pool of base learning ensembles for data stream $i$\\
$K$ & Base learning ensemble size\\
$h_i^{j,k}$& The $k$th sub-classifier of $H_i^{j}$\\
$\lambda_{h^{j,k}_{i}}^{s}$ & Sum of $h_{i}^{j,k}$'s probabilistic predictions for target examples that it correctly ($s = sc$) or incorrectly ($s = sw$) classifies\\
$\alpha_{h^{j,k}_{i}}$ & $h^{j,k}_{i}$'s predictive performance on current target concept\\
\hline
\end{tabular}}
\label{tab:notation}
\vspace{-0.4cm}
\end{table}

\section{Proposed Method}
\label{sec:ProposedMethod}



\begin{figure*}[tb]
\centerline{\includegraphics[width=.9\textwidth, height =7cm]{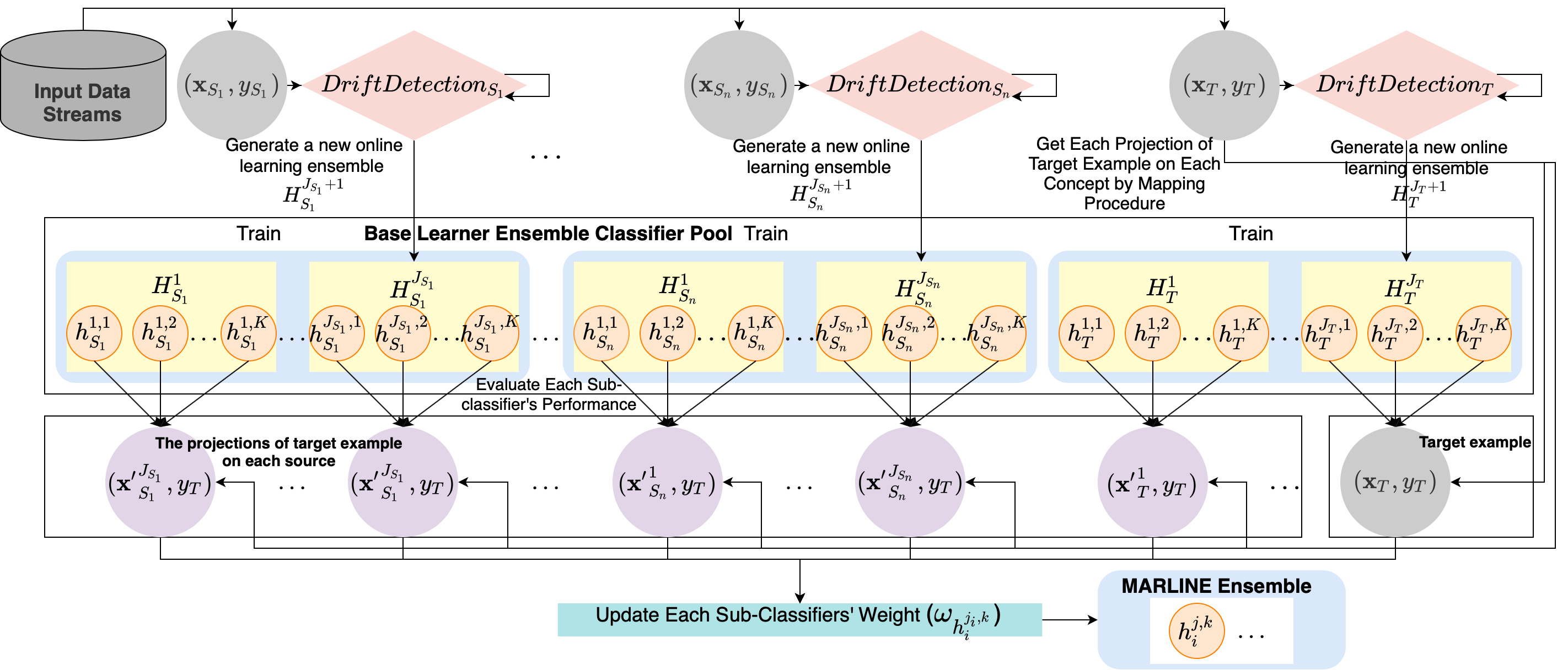}}
\caption{Overview of the proposed MARLINE training framework.}
\label{fig:frame}
\vspace{-0.4cm}
\end{figure*}
This section introduces MARLINE. An overview of MARLINE’s training framework is given in Figure \ref{fig:frame}, and its Java implementation is available in \cite{code}. MARLINE considers that we have multiple input data streams (represented in grey in the figure), where the concepts of the sources and target streams may be different. However, classifiers learned from the source concepts may still be used to improve the predictive performance in the target domain. For that, each concept observed from each data stream is learnt by an independent online learning ensemble, which we refer to as \textit{base learning ensemble} (shown in yellow). To identify different concepts, a concept drift detection method is used.
Whenever a new target example needs to be predicted, MARLINE uses a mapping between the current target concept and each source concept, so that the target example is geometrically projected onto the space of the source concept (the projections are shown in purple). Through this mapping, the target example can be predicted by different sub-classifiers (shown in orange) trained by different base learning ensembles. 
MARLINE weights each sub-classifier depending on how useful it is for predicting the projected target (shown in cyan-blue). The prediction by MARLINE is based on the weighted majority vote of the predictions of all the sub-classifiers that compose all the source and target ensembles. The set of all weighted sub-classifiers is referred to as the \textit{MARLINE ensemble}.


Section \ref{subsec:train} introduces MARLINE's training procedure. Section \ref{subsec:mapping} presents the mapping procedure for concept projections, which is undertaken with respect to the centroids of each concept. The centroids calculation is explained in Section \ref{subsec:updateCtd}. Section \ref{subsec:ensembleWeight} discusses the weighting of sub-classifiers that compose the MARLINE ensemble used for predictions.  Section \ref{subsec:vote} explains the voting procedure used for making predictions. Section \ref{subsec:time-complexity} shows the time complexity.

\subsection{Training}
\label{subsec:train}
\begin{algorithm}[!t]
\caption{Learning Procedure of MARLINE.}
\label{alg:learn_algorithm}
\textbf{Input}: Data streams $D_i,  \quad \forall i \in \{S_1,S_2, \cdots, S_n, T\}$\\ \label{lin:input}
\textbf{Parameter}: Online Base Learning Ensemble Algorithm; Base Learning Ensemble Size $K$; Drift Detection Method; Time Forgetting Factor $0 < \theta \leq 1$; Performance Index $0 \leq \sigma \leq 1$
\label{lin:parameter}
\vspace{-0.4cm}
\begin{algorithmic}[1] 
\STATE Set $H_i = \emptyset$, $J_i = 0$, $\mathcal{M} = \emptyset$; $\forall i \in \{S_1,S_2, \cdots, S_n, T\}$\label{lin:initial}
\WHILE{Receive a new example $(\textbf{x}_i,y_i)$, $\quad i \in \{S_1,S_2,\cdots,S_n,T\}$}
\IF {$i \notin \mathcal{M}$} \label{lin:checkM}
\STATE $\mathcal{M} \leftarrow \mathcal{M} \cup i$ \label{lin:addM}
\STATE $J_i \leftarrow 1$ (Number of online base learning ensembles associated to $i$ is 1) \label{lin:initialJ}
\STATE Initialise online base learning ensemble $H_i^{J_i}$ \label{lin:initialH}
\STATE $H_i \leftarrow H_i \cup H_i^{J_i}$ \label{lin:addH}
\STATE Calculate centroids $\textbf{c}_{i,J_i}$ as shown in Section \ref{subsec:updateCtd}, used to create the mapping as shown in Section \ref{subsec:mapping}\label{lin:initialCentroid}
\vspace{-0.4cm}
\STATE $\lambda_{h^{J_i,k}_{i}}^{sc}  \leftarrow 0, \lambda_{h^{J_i,k}_{i}}^{sw} \leftarrow 0, \alpha_{h^{J_i,k}_{i}} \leftarrow 1, \forall k \leq K$ \label{lin:intialIndexs}
\ENDIF
\IF {$\textit{DriftDetection}_i \left(\textbf{x}_i, y_i \right) = true$} \label{lin:detectDrift}
\STATE Initialise new online base learning ensemble $H_{i}^{J_i+1}$ \label{lin:initialNewH}
\STATE $J_i \leftarrow J_i + 1$ \label{lin:updateJ}
\STATE $H_i \leftarrow H_i \cup H_i^{J_i}$ \label{lin:updateH}
\STATE $\lambda_{h^{J_i,k}_{i}}^{sc} \leftarrow 0, \lambda_{h^{J_i,k}_{i}}^{sw} \leftarrow 0, \alpha_{h^{J_i,k}_{i}} \leftarrow 1,\forall k \leq K$ \label{lin:intialUpdateParameter}
\IF {$i = T$} \label{lin:resetParameterBeg}
\STATE $\lambda_{h^{j,k}_{i'}}^{sc}  \leftarrow 0, \lambda_{h^{j,k}_{i'}}^{sw} \leftarrow 0, \alpha_{h^{j,k}_{i'}} \leftarrow 1, \forall i' \in \mathcal{M}$, $j \leq J_{i'}$, $k \leq K$\\
\ENDIF \label{lin:resetParameterEnd}
\ENDIF
\STATE OnlineBaseLearnerAlgorithm$ \{H_{i}^{J_i},  \left(\textbf{x}_i, y_i \right) \}$ \label{lin:classifierLearn}
\STATE Update centroids $\textbf{c}_{i,J_i}$ \label{lin:updateCentroid} as shown in Section \ref{subsec:updateCtd}, used to create the mapping as shown in Section \ref{subsec:mapping}
\IF {$i=T$}
\STATE Update each sub-classifier's weight as shown in Section \ref{subsec:ensembleWeight} \label{lin:checkT}
\ENDIF
\ENDWHILE
\end{algorithmic}
\end{algorithm}

The pseudo-code of MARLINE's training process is shown in Algorithm \ref{alg:learn_algorithm}.
The set $\mathcal{M} \subseteq  \{S_1,\cdots,S_n,T\}$ is the set of all the sources and target for which an online base learning ensemble has already been generated.
When a new example $(\textbf{x}_i, y_i)$ is received from a source or target data stream $i$ for the first time, the proposed method creates one online base learning ensemble $H_i^1$ for this source or target (lines \ref{lin:checkM} to \ref{lin:initialH}). Any online learning ensemble method can be used, e.g., online boosting or bagging \cite{oza2005online}. Ensembles are used here because their diversity increases the chances that at least some of the sub-classifiers become useful for predicting the target \cite{du2019multi}.

As the concept of each data stream $i$ may change due to concept drifts, each source/target $i$ is associated with a pool of base learning ensembles $H_i$, where each ensemble may represent a different concept observed from that source/target.
Each ensemble $H_i^{j}$ within the pool contains $K$ sub-classifiers $h_i^{j,k}$, where $1 \leq k \leq K$. The newly created ensemble $H_i^1$ is added to its corresponding ensemble pool $H_i$ (line \ref{lin:addH}). This pool receives additional ensembles when $i$ suffers from concept drifts, as explained later in this section.
After having added $H_i^1$ to $H_i$, the method discussed in Section \ref{subsec:updateCtd} is used to calculate the centroid associated with the concept $p^1_i$ (line \ref{lin:initialCentroid}). This centroid is used to create the mapping from the target to the source concepts, as explained in Section \ref{subsec:mapping}. Line \ref{lin:intialIndexs} is used to initialise the weights associated with each sub-classifier. These weights are used to identify which sub-classifiers are more important for predicting the projected target, and their calculation is explained in Section \ref{subsec:ensembleWeight}. 

Whenever a new training example $(\textbf{x}_i, y_i)$ is received from stream $i$, MARLINE runs a drift detection method on $i$. Any drift detection method could be used, e.g., HDDM \cite{frias2014online}. If the drift detection method requires monitoring a predictive model representing $i$, the most recent ensemble $H_i^{J_i}$ is used. If a concept drift is detected, the proposed method creates a new online base learning ensemble $H_i^{J_i+1}$ with the initialisation of its weights and connection with the pool of ensembles $H_i$ (lines \ref{lin:detectDrift} to \ref{lin:intialUpdateParameter}).
If the new example belongs to the target domain, all the values of $\lambda_{h^{j,k}_{i'}}^{sc}, \lambda_{h^{j,k}_{i'}}^{sw}, \alpha_{h^{j,k}_{i'}}, \forall i' \in \mathcal{M}$, $j \leq J_{i'}$, $k \leq K$ for all the sub-classifiers are reset (lines \ref{lin:resetParameterBeg} to  \ref{lin:resetParameterEnd}) so that all the sub-classifiers' weights can start adapting to the new target distribution.

After checking for concept drift, the most recent ensemble $H_i^{J_i}$ created for the source or target $i$ is trained on the current example ($\textbf{x}_i,y_i$) (line \ref{lin:classifierLearn}). The centroid of concept $p^{J_i}_i$ is updated (line \ref{lin:updateCentroid}), as explained in Section \ref{subsec:updateCtd}.

If the new training example belongs to the target stream, the sub-classifiers' weighting scheme discussed in Section \ref{subsec:ensembleWeight} is applied (line \ref{lin:checkT}). The mapping procedure shown in Section \ref{subsec:mapping} is used in the weighting scheme.

\subsection{Mapping Procedure}
\label{subsec:mapping}
The main purpose of the mapping procedure is to create the projection $(\textbf{x$'$}_i^{j})$ of the current target example $(\textbf{x}_T)$ on the source concept $p^j_i$.
Therefore, given a current target example, the classifiers trained with a certain source concept can make a prediction on the projection of this target example based on the knowledge learned from the source concept.
After a concept drift has been detected in the target data stream, any previous concept in that stream is also regarded as a source concept. From this point onward, we will use the term ``source+'' instead of ``source'' when the past target concepts are included as sources. Therefore, $i$ and $j$ in the source+ concept $p^j_i$ are defined as follows:  $\{i,j\ | \ i \in \{S_1, \cdots, S_n\}, 0 < j \leq J_i\} \cup \{i,j \ | \ i = T, 0 < j < J_i\}$.

To seek the projection $(\textbf{x$'$}_i^{j})$ of target example $(\textbf{x}_T)$ on $p^j_i$, a mapping function between $p^j_i$ and $p^{J_T}_T$ is required. However, for online learning we only store the latest example in memory over time. To build the mapping function between the source+ and target concepts without retrieving the overall historical data, we propose the following procedure. Consider a pair of reference points in a given source+ concept, and another pair in the target concept. For instance, for a single concept $p_i^j$, the pair of reference points can be the centroids of the distributions $p_i^j(\textbf{x}|y), y \in [-1,+1]$, i.e.:
\begin{equation}
    \textbf{c}^y_{i,j} = [c^1, c^2, \cdots, c^d]; y\in [-1,+1]
\end{equation}
\noindent The calculation of $\textbf{c}^y_{i,j}$ is shown in Section \ref{subsec:updateCtd}.

We connect the pairs of reference points of a given concept using a vector 
$\overrightarrow{V_{i,j}} = \textbf{c}^{y=1}_{i,j} - \textbf{c}^{y=-1}_{i,j}$.
%
The transformation matrix $R$ between $\overrightarrow{V_{i,j}}$ and $\overrightarrow{V_{T,J_T}}$ can be considered as the mapping function of any two vectors between $p^j_i$ and $p^{J_T}_T$:
\begin{equation}
\overrightarrow{V_{i,j}} = R \cdot \overrightarrow{V_{T,J_T}}
\end{equation}

\noindent Therefore, the source+ vector $\overrightarrow{V_{i,j}}$ can be seen as a projection of the target vector $\overrightarrow{V_{T,J_T}}$ on $p^j_i$.

The transformation matrix $R$ can be calculated by Eqs. (\ref{eq.R_start})-(\ref{eq.R_end}), where $I_d$ is the identity matrix with $d$ dimensions.

\vspace{-0.3cm}
\begin{gather}
\label{eq.R_start}
\overrightarrow{u} = \frac{\overrightarrow{V_{i,j}}\inv}{\norm{{\overrightarrow{V_{i,j}}\inv}}}\hspace{0.1cm},\hspace{0.4cm}
\overrightarrow{v} =
\frac{\overrightarrow{V_{T,J_T}}\inv}{\norm{\overrightarrow{V_{T,J_T}}\inv}}\\
A =  I_d - (\overrightarrow{u}+\overrightarrow{v}) \cdot 2 \cdot \frac{(\overrightarrow{u} + \overrightarrow{v})\inv \cdot I_d}{(\overrightarrow{u} + \overrightarrow{v})\inv \cdot (\overrightarrow{u} + \overrightarrow{v})}\\
R = (A - \overrightarrow{v} \cdot 2 \cdot \frac{\overrightarrow{v}\inv \cdot A}{\overrightarrow{v}\inv \cdot \overrightarrow{v}} )\cdot \frac{\norm{\overrightarrow{V_{i,j}}}}{\norm{\overrightarrow{V_{T,J_T}}}} \label{eq.R_end}
\end{gather}

When a new example $(\textbf{x}_T)$ from the target domain is received, the vector between the new example and one centroid $(\textbf{c}^{y=1}_{T,J_T})$ of the target concept can be computed as follows:
\begin{equation}
    \overrightarrow{VT} = \textbf{x}_T - \textbf{c}^{y=1}_{T,J_T}
\end{equation}

The projection of $\overrightarrow{VT}$ on $p^j_i$ can be calculated as follows:
\begin{equation}
   \overrightarrow{VI_j} = R \cdot \overrightarrow{VT}
\end{equation}

The projection $(\textbf{x$'$}_i^{j})$ is the sum of $\textbf{c}^{y=1}_{i,j}$ and $\overrightarrow{VI_j}$:
\begin{align}
    \textbf{x$'$}_i^{j} = \textbf{c}^{y=1}_{i,j} + \overrightarrow{VI_j}
\end{align}

\subsection{Calculating and Updating the Centroids}
\label{subsec:updateCtd}
For saving computational time, we update the centroids of each concept instead of updating transformation matrix $R$ at each time step. The transformation matrix $R$ will be calculated whenever a target example needs to be predicted.

The centroids of the concept $p_i^j(\textbf{x}, y)$ are dynamically updated based on examples $(\textbf{x}_i,y_i)$ received from data stream $i$ during the time window since the concept $j$ has become active. 
During this time window, if no example with label $y_i$ has been seen before, the centroid $\textbf{c}^{y=y_i}_{i,J_i}$ is set as $\textbf{x}_i$. Otherwise, it is updated as follows:
\begin{gather}
    \textbf{sumC}_{i, J_i}^{y=y_i} = \theta \textbf{sumC}_{i, J_i}^{y=y_i} + \textbf{x}_i\\
    \textbf{c}_{i, J_i}^{y=y_i} = \frac{\textbf{sumC}_{i, J_i}^{y=y_i}}{\sum_{t' = 1}^{L} \theta^{(t'-1)}}
\end{gather}

\noindent where $L$ is the number of the training examples received in the data stream $i$ since the concept $p_i^j$ has become active, and $\theta, 0 < \theta \leq 1$, is a pre-defined forgetting factor used to reduce the weight given to the historical examples. It helps to deal with non-stationary environments. The summation in the denominator is a normalisation factor, which can be updated in an online manner.



\subsection{Sub-Classifiers Weighting}
\label{subsec:ensembleWeight}
When an example $(\textbf{x}_T, y_T)$ from the target domain is received, all the sub-classifiers' weights $\omega_{h_i^{j,k}}$ are updated. We assign larger weights to the sub-classifiers which focus on \textit{harder} classified examples. The sub-classifier's weight $\omega_{h_i^{j,k}}$ depends on the corresponding sub-classifier's performance $\alpha_{h_i^{j,k}}$ on the current target concept $p_T^{J_T}$. All the sub-classifiers' performances $\alpha_{h_i^{j,k}}, \forall i \in \mathcal{M}, 0 < j \leq J_i, 0 < k \leq K$ are updated based on the corresponding projection $(\textbf{x$'$}_i^{j}, y_T)$ of the current target example $(\textbf{x}_T, y_T)$, which has been obtained by the mapping procedure explained in Section \ref{subsec:mapping}. When $i = T, j = J_T$, the projection $(\textbf{x$'$}_i^{j}, y_T)$ is set to the current target example $(\textbf{x}_T, y_T)$.

The sub-classifiers' performance is initialised with $\alpha_{h_i^{j,k}} = 1, \forall i \in \mathcal{M}, 0 < j \leq J_i, 0 < k \leq K$. To update each sub-classifier's performance, the current example's weight needs to be calculated. We get the prediction of each sub-classifier on the corresponding target example projection. The example's weight $\frac{SW}{SC}$ can be calculated as follows:
\begin{gather}
    \label{eq:wsc}
    SC \leftarrow \sum_{{i}\in\mathcal{M}}\sum_{{j}=1}^{J_{i}}\sum_{k=1}^{K} \alpha_{h^{j,k}_{i}}P(h^{j,k}_{i}(\textbf{x$'$}_{i}^{j})= y_T)\\ 
    SW \leftarrow \sum_{{i}\in\mathcal{M}}\sum_{{j}=1}^{J_{i}}\sum_{k=1}^{K} \alpha_{h^{j,k}_{i}}P(h^{j,k}_{i}(\textbf{x$'$}_{i}^{j})\neq y_T) \label{eq:wsw}
\end{gather}

\noindent where $P(h(\textbf{x}) = y)$ is the probability of class $y$, estimated by the sub-classifier $h(\textbf{x})$, and $\alpha_{h^{j,k}_{i}}$ is the performance computed based on the target examples received \textit{before} receiving $(\textbf{x}_T,y_T)$. The example's weight $\frac{SW}{SC}$ is used to indicate how confident the MARLINE ensemble is for the current example. We expect that the more confident the MARLINE ensemble is, the smaller weight the example receives, and vice versa.

Consider that $\lambda_{h^{j,k}_{i}}^{sc}$ is the sum of the probabilistic predictions of sub-classifier $h_{i}^{j,k}$ for the target examples that it correctly classifies and $\lambda_{h^{j,k}_{i}}^{sw}$ is the sum of the probabilistic predictions for misclassified target examples. The performance $\alpha_{h^{j,k}_{i}}$ of each sub-classifier can be computed incrementally by:
\begin{gather}
    \label{eq.alpha_start}
    \lambda_{h^{j,k}_{i}}^{sc} \leftarrow \theta\lambda_{h^{j,k}_{i}}^{sc} + \frac{SW}{SC}\frac{\alpha_{h^{j,k}_{i}}P(h^{j,k}_{i}(\textbf{x$'$}_{i}^{j})= y_T)}{SC}\\ 
    \lambda_{h^{j,k}_{i}}^{sw} \leftarrow \theta\lambda_{h^{j,k}_{i}}^{sw} + \frac{SW}{SC}\frac{\alpha_{h^{j,k}_{i}}P(h^{j,k}_{i}(\textbf{x$'$}_{i}^{j})\neq y_T)}{SW}\\ 
    \alpha_{h^{j,k}_{i}} \leftarrow \frac{\lambda_{h^{j,k}_{i}}^{sc}}{\lambda_{h^{j,k}_{i}}^{sc} + \lambda_{h^{j,k}_{i}}^{sw}}, \label{eq.alpha_end}
\end{gather}

\vspace{-0.2cm}
\noindent where $0 < \theta \leq 1$ is a forgetting factor and $\frac{\alpha_{h^{j,k}_{i}}P(h^{j,k}_{i}(\textbf{x$'$}_{i}^{j})= y_T)}{SC}$ represents how much contribution sub-classifier $h^{j,k}_{i}$ makes in the MARLINE ensemble to vote for the projection $(\textbf{x$'$}_{i}^{j})$ of $(\textbf{x}_T)$ with label $y_T$. Thus, $\alpha_{h^{j,k}_{i}}$ is the current performance percentage of each sub-classifier, giving more focus to more recently arriving target examples.

The weights of all the sub-classifiers associated with $\alpha_{h^{j,k}_{i}} > \sigma$ are assigned to their predictive performance $\alpha_{h^{j,k}_{i}}$ normalised by the sum of the predictive performances of all the sub-classifiers associated with $\alpha_{h^{j,k}_{i}} > \sigma$. The weight $\omega_{h^{j,k}_{i}}$ can be formulated as follows:
\begin{equation}
    \omega_{h^{j,k}_{i}} =\\
\resizebox{0.85\linewidth}{!}{$\begin{cases}
\frac{1}{\sum_{{i'}\in\mathcal{M}}\sum_{{j'}=1}^{J_{i'}}\sum_{k=1}^{K} (\alpha_{h^{j',k}_{i'}}>\sigma ? \ \alpha_{h^{j',k}_{i'}}:0)}\alpha_{h^{j,k}_{i}},& \alpha_{h^{j,k}_{i}} > \sigma \\
0, & otherwise
\end{cases}$} \label{eq:omega}
\end{equation}

\noindent where $\sigma$ is a pre-defined parameter, (testCondition ? v1 : v2) retrieves v1 if testCondition is true, and v2 otherwise. 

\subsection{Voting Procedure for Making Predictions}
\label{subsec:vote}
When a prediction is needed for a target instance $(\textbf{x}_T)$, we multiply the corresponding weights of the sub-classifiers with the probabilistic prediction made by each sub-classifier on their corresponding projection $(\textbf{x}_{i'}^j)$ of the current target example. All sub-classifiers $h^{j,k}_{i'}$, ${i'}\in \mathcal{M}$, $j \in \{1,\cdots, J_i\}$, $k \in \{1,\cdots,K\}$ are considered for this purpose. Afterwards, we obtain the sum of the weighted prediction probabilities of all classes and use majority vote to decide the predicted class.

\subsection{Time Complexity Analysis}
\label{subsec:time-complexity}

When learning a target training example, MARLINE's training time complexity is 
$\mathcal{O}(f_{DD} + f_H + (J_{S_1}+J_{S_2}+\cdots +J_{S_n} + J_T) d^2 + (J_{S_1}+J_{S_2}+\cdots +J_{S_n} + J_T) K \times f_h)$, where $f_H$, $f_{DD}$ and $f_h$ are the time complexities for training the base learning ensemble with the example, running the drift detection method and getting the prediction from a sub-classifier.

When learning a source training example, MARLINE's training time complexity is $\mathcal{O}(f_{DD} + f_H + d)$.

MARLINE's time complexity for prediction is $\mathcal{O}((J_{S_1}+J_{S_2}+\cdots +J_{S_n} + J_T)d^2 + (J_{S_1}+J_{S_2}+\cdots +J_{S_n}) K \times f_h)$

Details on the complexity estimation can be found in the Supplementary Material of this paper \cite{supplementary}. 

\section{Experiments Setup}
\label{sec:Experiments}

We evaluate MARLINE under several different conditions, including stationary environments, non-stationary environments with different types of concept drifts, and different target data stream sizes. 
Artificial datasets enable us to better understand \textit{when} and \textit{how} MARLINE can be helpful. Real world datasets enable us to check whether MARLINE can work well in practice.

\subsection{Datasets}

We use the same three artificial datasets of similar target and sources as those of \cite{du2019multi} and generated additional datasets where the source was non-similar to the targets. These datasets have two numeric features and one binary output. The examples belonging to each output class were generated by a Gaussian distribution as shown in Table \ref{tab:artDistributions}, where 
each dataset is composed of several (target and sources) data streams. The three datasets with similar sources use only the target and similar source data streams from Table \ref{tab:artDistributions}, whereas the three datasets with non-similar source use only the target and non-similar source data stream from Table \ref{tab:artDistributions}.
The datasets simulate a stationary environment (no drift)
and two types of non-stationary environments (abrupt drift and incremental drift) on the target stream. Each dataset also has three different versions based on different class size scenarios (small, medium, large) by varying the number of target training examples of each class in {50, 500, 5000}. Each source stream has 5000 training examples of each class without any concept drift.

\begin{table}[b]
\vspace{-0.45cm}
\caption{The parameters of the Gaussians of Artificial Datasets.}
\vspace{-0.2cm}
\scalebox{0.67}{
\begin{tabular}{|l|l|l|l|l|l|}
\hline
Datasets                                    & Domain Type              & \multicolumn{2}{l|}{Data Stream}       & Class 0 Centre & Class 1 Centre \\ \hline
\multirow{3}{*}{No Drift Datasets}          & Target                   & \multicolumn{2}{l|}{Target}            & (2,3)          & (7,8)          \\ \cline{2-6} 
                                            & Similar                  & \multicolumn{2}{l|}{Source}            & (2,1)          & (7,8)          \\ \cline{2-6} 
                                            & Non-Similar              & \multicolumn{2}{l|}{Source}            & (-2,-3)        & (-7,2)         \\ \hline
\multirow{4}{*}{Abrupt Drift Datasets}      & \multirow{2}{*}{Target}  & \multirow{2}{*}{Target} & Before Drift & (2,3)          & (7,8)          \\ \cline{4-6} 
                                            &                          &                         & After Drift  & (2,9)          & (5,4)          \\ \cline{2-6} 
                                            & Similar                  & \multicolumn{2}{l|}{Source}            & (2,9)          & (5,4)          \\ \cline{2-6} 
                                            & Non-Similar              & \multicolumn{2}{l|}{Source}            & (-2,-3)        & (-7,2)         \\ \hline
\multirow{9}{*}{Incremental Drift Datasets} & \multirow{2}{*}{Target}  & \multirow{2}{*}{Target} & Before Drift & (2,3)          & (7,8)          \\ \cline{4-6} 
                                            &                          &                         & After Drift  & (2,3)          & (7,8)          \\ \cline{2-6} 
                                            & \multirow{6}{*}{Similar} & \multicolumn{2}{l|}{Source 1}          & (2,3)          & (7,8)          \\ \cline{3-6} 
                                            &                          & \multicolumn{2}{l|}{Source 2}          & (3,4)          & (6,7)          \\ \cline{3-6} 
                                            &                          & \multicolumn{2}{l|}{Source 3}          & (4,5)          & (5,6)          \\ \cline{3-6} 
                                            &                          & \multicolumn{2}{l|}{Source 4}          & (5,6)          & (4,5)          \\ \cline{3-6} 
                                            &                          & \multicolumn{2}{l|}{Source 5}          & (6,7)          & (3,4)          \\ \cline{3-6} 
                                            &                          & \multicolumn{2}{l|}{Source 6}          & (7,8)          & (2,3)          \\ \cline{2-6} 
                                            & Non-Similar              & \multicolumn{2}{l|}{Source}            & (-2,-3)        & (-7,2)         \\ \hline

\end{tabular}}
\begin{tablenotes}
\item  The covariance matrix for each class of each domain is $\begin{pmatrix}
1 & 0\\
0 & 2
\end{pmatrix}$ except for target in the no drift and non-similar datasets, which uses $\begin{pmatrix}
2 & 0\\
0 & 2
\end{pmatrix}$ for both classes. For the incremental drift, the centres of the Gaussian of class 0 and 1 move towards each other by one unit at each 100, 1000 and 10000 time steps for the datasets with class sizes of 50, 500,  5000, respectively, until the Gaussians of class 0 and 1 swap location. This leads to intermediate concepts that are equivalent to each of the similar sources 1 to 6.
\end{tablenotes}
\label{tab:artDistributions}
\end{table}

The real-world datasets are acquired from London bike sharing dataset \cite{LondonBike:2019} and Bike Sharing in Washington D.C. dataset \cite{fanaee2014event}. The task is to classify whether rental bikes are in low or high demand. We use the median of the total count of the rental bikes of a given dataset to indicate low and high demands in this dataset. We select the features shared by the two datasets (actual temperature, feeling temperature, humidity and wind speeds) to unify the feature dimension. Each dataset is divided into three sub-datasets based on holiday, weekend and weekday and compose the following three scenarios: We choose weekdays from Washington D.C. as the source, and (1) holidays and (2) weekends in London are the targets. We make weekends in Washington D.C. as the source and (3) weekdays in London are the target. The aim of these three different target sub-datasets is to create small, medium and large stream sizes (Holiday: 384, Weekend: 4970, Weekday: 12060). 

\subsection{Benchmark Methods and Evaluation Measures}
MARLINE was compared against Melanie \cite{du2019multi}, Adaptive Random Forest \cite{gomes2017adaptive}, Dynamic Weighted Majority (DWM) \cite{kolter2007dynamic}, Online Bagging\cite{oza2005online}, Online Boosting \cite{oza2005online}, Online Bagging with Drift Detection and Online Boosting with Drift Detection. Melanie was chosen for the comparison because it is the state-of-the-art multi-source transfer learning approach for non-stationary data streams. As MSCRDR \cite{dong2019multistream} and COMC \cite{tao2019comc}, Melanie is only able benefit from source concepts when they share the same task as the target concept. However, different from these approaches, Melanie has the advantage of being able to detect when source tasks are dissimilar to the target, avoiding to hinder predictive performance on the target when that is the case. Comparing MARLINE against Melanie will reveal whether MARLINE's mapping function is helpful to improve predictive performance against an approach that is only able to benefit from source concepts when they are similar to the target concept. 

Adaptive Random Forest and DWM were chosen because they are popular data stream learning approaches available in the Massive Online Analysis (MOA) tool \cite{bifet2010moa}. Comparing against them shows whether MARLINE can outperform the popular approaches. Online bagging and online boosting \cite{oza2005online} were included as baseline ensemble approaches with no strategy to deal with concept drift. They provide a desired lower bound for the predictive performance achieved by MARLINE and any other approaches for non-stationary environments. Additionally, they were also applied in combination with a drift detection method, which resets the models upon drift alarm, to enable these approaches to cope with drifts. Comparing against the combination shows whether or not MARLINE is able to benefit from sources in general.  

MARLINE without source data streams was also used in the comparison because mapping is also performed between the old target concepts and the current target concept. Including this approach shows whether or not it is beneficial to use different sources especially for the initial learning stage, rather than only mapping between old and new target concepts.

Both MARLINE and Melanie have been investigated with online bagging and online boosting as base learning ensemble methods \cite{oza2005online}. All ensemble approaches used Hoeffding trees \cite{domingos2000mining} as the basic units of learning, except for one of the compared approaches (ARF), which is based on ARFHoeffding Tree \cite{gomes2017adaptive}. 
Two drift detection methods (DDM \cite{gama2004learning} and $HDDM_A$ \cite{frias2014online}) were used for all the approaches that require drift detection. DDM is a well known method.
$HDDM_A$ has been recently shown to perform well compared to other drift detection methods when configuring ensembles \cite{de2019overview}.

Thirty runs were performed for all the compared approaches, except for DWM \cite{kolter2007dynamic}, which is deterministic and requires a single run. The average accuracy across the 30 runs is reported.

Grid search was used to tune the hyperparameters of each  approach on each dataset based on a preliminary run.
For Online Bagging and Boosting, the sizes of the sub-classifiers varied in 1:1:30. For DWM, $\beta$ varied in 0:0.1:1, period $p$ = 1, and the weight threshold for removing sub-classifiers was 0.01. For ARF, the number of trees varied in 10:1:30 (MOA restricts the minimum ARF ensemble size to 10). For Melanie, the forgetting factor varied in $0.9:0.01:1$. For MARLINE, $\theta = 0.9:0.01:1$ and $\sigma = 0.1:0.1:1$. The grid searches' results are in the Supplementary Material of this paper \cite{supplementary}. 

For the artificial data streams, the predictive performance is calculated prequentially and reset upon the real location of the drifts \cite{minku2011ddd}. In the artificial datasets, we know exactly when the concept drifts happen. This evaluation framework will reset the accuracy to zero when the concept drifts (or the increments of an incremental drift) occur. This enables us to measure the performance on each concept separately without being affected by the previous concepts.
For real world data streams, the predictive performance of all the approaches is evaluated using sliding windows \cite{gama2009issues} with the size of 10\% of the target data stream. Friedman and their Nemenyi post-hoc tests were used to compare the predictive performance of all approaches, on each dataset.

\section{Experiment Results}

\begin{figure}[t]
\begin{subfigure}{0.235\textwidth}
\centerline{\includegraphics[scale=0.22]{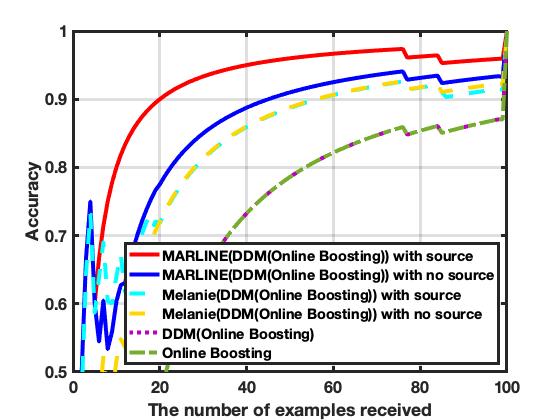}}
\caption{No Drift; class size of 50}
\label{subfig:e1_50_ns}
\end{subfigure}
\begin{subfigure}{0.235\textwidth}
\centerline{\includegraphics[scale=0.22]{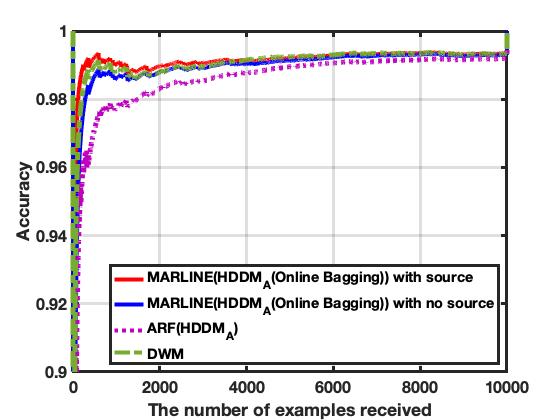}}
\caption{No Drift; class size of 5000}
\label{subfig:e1_5000_ns}
\end{subfigure}
\begin{subfigure}{0.235\textwidth}
\centerline{\includegraphics[scale=0.22]{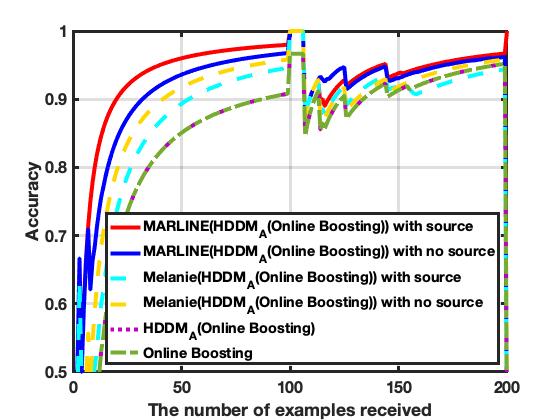}}
\caption{Abrupt; class size of 50}
\label{subfig:e2s_50_ns}
\end{subfigure}
\begin{subfigure}{0.235\textwidth}
\centerline{\includegraphics[scale=0.22]{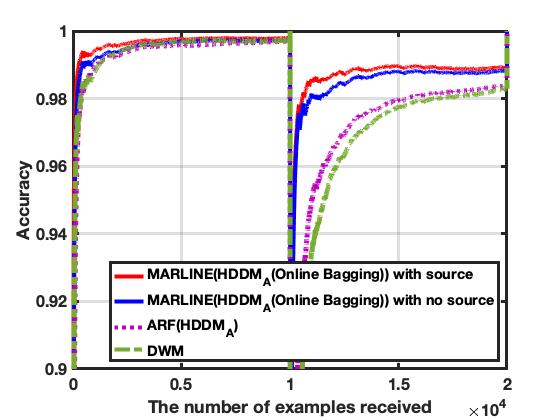}}
\caption{Abrupt; class size of 5000}
\label{subfig:e2s_5000_ns}
\end{subfigure}
\begin{subfigure}{0.235\textwidth}
\centerline{\includegraphics[scale=0.22]{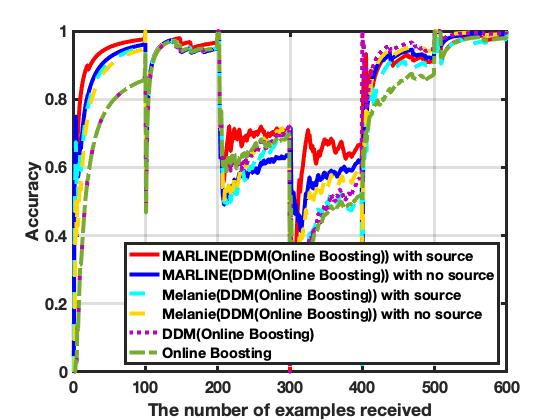}}
\caption{Incremental; class size of 50}
\label{subfig:e2i_50_ns}
\end{subfigure}
\begin{subfigure}{0.235\textwidth}
\centerline{\includegraphics[scale=0.22]{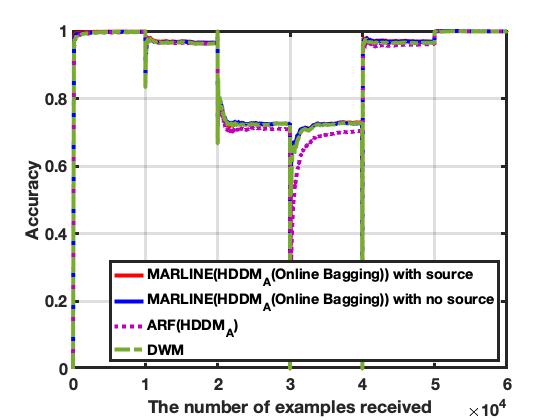}}
\caption{Incremental; class size of 5000}
\label{subfig:e2i_5000_ns}
\end{subfigure}
\caption{Average Accuracy with Non-Similar Source.}
\label{fig: NS}
\vspace{-0.45cm}
\end{figure}


\begin{figure}[tb]
\begin{subfigure}{0.235\textwidth}
\centerline{\includegraphics[scale=0.22]{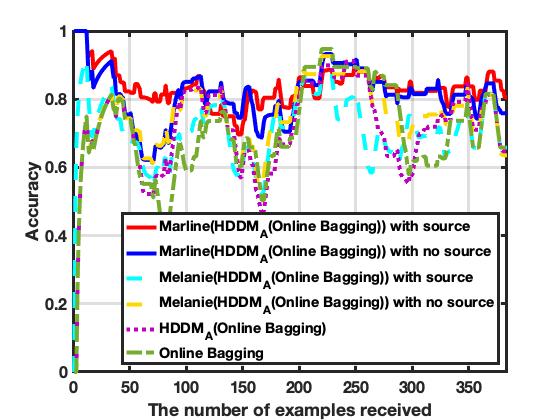}}
\caption{Holiday}
\label{subfig:holiday}
\end{subfigure}
\begin{subfigure}{0.235\textwidth}
\centerline{\includegraphics[scale=0.22]{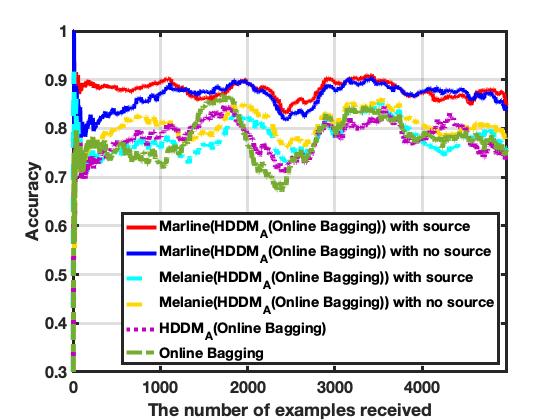}}
\caption{Weekend}
\label{subfig:weekend}
\end{subfigure}
\begin{subfigure}{0.235\textwidth}
\centerline{\includegraphics[scale=0.22]{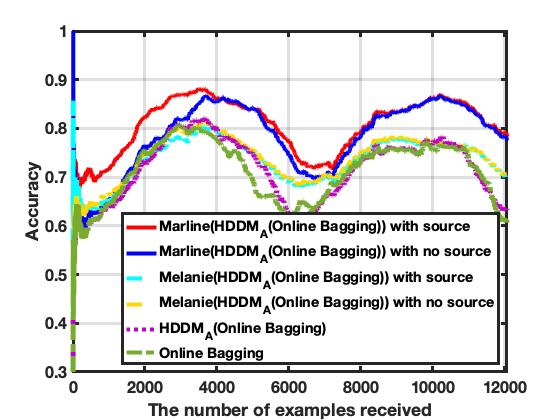}}
\caption{Weekday} 
\label{subfig:weekday_0.9}
\end{subfigure}
\begin{subfigure}{0.235\textwidth}
\centerline{\includegraphics[scale=0.22]{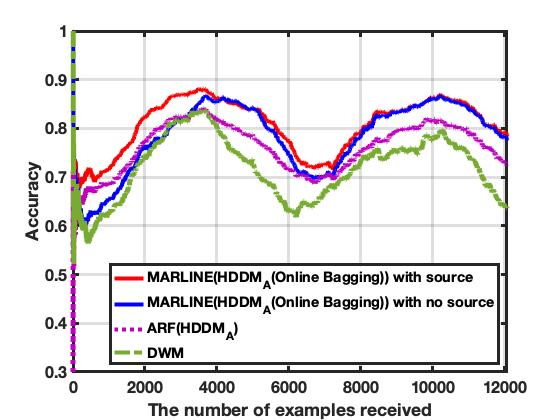}}
\caption{Weekday}
\label{subfig:weekday_other}
\end{subfigure}
\caption{Accuracy on Real World Datasets.}
\label{fig: Real}
\vspace{-0.45cm}
\end{figure}

\begin{table*}[tb]\centering
\caption{Friedman Ranks on Each Dataset.}
\vspace{-0.2cm}
\scalebox{0.625}{
\begin{tabular}{|l V{8} c|c|c|c|c|c|c|c|c V{8} c|c|c|c|c|c|c|c|c V{8} c|c|c|}
\hline
\multirow{2}{*}{$Dataset$}
&\multicolumn{9}{c V{8}}{Non-Similar Source}
&\multicolumn{9}{c V{8}}{Similar Source}
&\multicolumn{3}{c|}{Real-World Data}\\
\cline{2-22}
&\multicolumn{3}{c|}{No Drift}
&\multicolumn{3}{c|}{Abrupt}
&\multicolumn{3}{c V{8}}{Incremental}
&\multicolumn{3}{c|}{No Drift}
&\multicolumn{3}{c|}{Abrupt}
&\multicolumn{3}{c V{8}}{Incremental}
&{Holiday}
&{Weekend}
&{Weekday}
\\
\hline
Class Size or Target Stream Size
&50&500&5000
&50&500&5000
&50&500&5000
&50&500&5000
&50&500&5000
&50&500&5000
&384&4970&12060
\\
\hline
\textbf{Marline(DDM(Online Bagging)) with source}&
\cellcolor[gray]{0.9}\textbf{\textcolor{red}{2.8}}&\cellcolor[gray]{0.9}\textbf{2.3}&2.7&5.9&6.4&2.7&10.9&9.5&14.2&7.9&\cellcolor[gray]{0.9}\textbf{3.0}&3.5&\cellcolor[gray]{0.9}\textbf{3.5}&8.4&7.9&10.5&9.6&13.2&\cellcolor[gray]{0.9}\textbf{6.8}&4.9&2.0\\
\textbf{Marline(DDM(Online Boosting)) with source}&
\cellcolor[gray]{0.9}\textbf{\textcolor{red}{2.8}}&11.0&19.9&\cellcolor[gray]{0.9}\textbf{3.2}&\cellcolor[gray]{0.9}\textbf{\textcolor{red}{1.5}}&9.9&\cellcolor[gray]{0.9}\textbf{8.5}&14.2&16.2&\cellcolor[gray]{0.9}\textbf{6.2}&7.3&8.5&6.1&3.5&7.9&10.6&12.5&15.3&10.0&13.6&8.3\\
\textbf{Marline($HDDM_A$(Online Bagging)) with source}&
\cellcolor[gray]{0.9}\textbf{\textcolor{red}{2.8}}&\cellcolor[gray]{0.9}\textbf{\textcolor{red}{1.2}}&\cellcolor[gray]{0.9}\textbf{\textcolor{red}{1.5}}&5.5&9.0&\cellcolor[gray]{0.9}\textbf{\textcolor{red}{1.4}}&11.1&5.3&\cellcolor[gray]{0.9}\textbf{\textcolor{red}{3.1}}&\cellcolor[gray]{0.9}\textbf{6.2}&3.8&3.5&\cellcolor[gray]{0.9}\textbf{\textcolor{red}{3.4}}&6.3&5.4&7.6&6.0&\cellcolor[gray]{0.9}\textbf{\textcolor{red}{4.3}}&\cellcolor[gray]{0.9}\textbf{\textcolor{red}{5.6}}&\cellcolor[gray]{0.9}\textbf{\textcolor{red}{1.6}}&\cellcolor[gray]{0.9}\textbf{\textcolor{red}{1.4}}\\
\textbf{Marline($HDDM_A$(Online Boosting)) with source}&
\cellcolor[gray]{0.9}\textbf{\textcolor{red}{2.8}}&10.2&13.8&\cellcolor[gray]{0.9}\textbf{\textcolor{red}{2.7}}&\cellcolor[gray]{0.9}\textbf{1.7}&4.4&\cellcolor[gray]{0.9}\textbf{\textcolor{red}{7.3}}&11.0&11.4&\cellcolor[gray]{0.9}\textbf{6.4}&8.7&9.2&\cellcolor[gray]{0.9}\textbf{5.0}&\cellcolor[gray]{0.9}\textbf{\textcolor{red}{2.7}}&\cellcolor[gray]{0.9}\textbf{\textcolor{red}{4.6}}&11.0&10.1&10.1&10.2&7.6&7.0\\
\textbf{Marline(DDM(Online Bagging)) without source}&
7.3&6.7&5.0&10.2&17.1&7.5&9.8&8.2&13.0&
11.0&9.5&7.5&13.0&17.8&8.6&12.2&12.0&16.2&\cellcolor[gray]{0.9}\textbf{7.5}&5.1&9.1\\
\textbf{Marline(DDM(Online Boosting)) without source}&
\cellcolor[gray]{0.9}\textbf{5.8}&16.4&14.4&5.8&3.9&12.3&12.6&10.5&14.5&9.6&17.5&16.4&9.4&5.9&13.5&15.3&14.3&17.5&6.7&10.0&20.1\\
\textbf{Marline($HDDM_A$(Online Bagging)) without source}&
10.0&4.7&5.0&10.8&23.2&8.6&9.0&\cellcolor[gray]{0.9}\textbf{\textcolor{red}{3.6}}&3.9&13.3&8.0&7.5&13.6&23.1&9.9&11.4&6.2&5.3&\cellcolor[gray]{0.9}\textbf{6.7}&2.5&4.6\\
\textbf{Marline($HDDM_A$(Online Boosting)) without source}&
\cellcolor[gray]{0.9}\textbf{5.8}&16.6&15.2&\cellcolor[gray]{0.9}\textbf{4.3}&8.5&10.6&12.5&10.8&12.1&9.6&17.6&17.1&7.7&9.2&12.0&14.9&14.1&14.6&\cellcolor[gray]{0.9}\textbf{6.5}&6.3&10.9\\
{Melanie(DDM(Online Bagging)) with source}&
19.6&8.0&8.0&16.7&13.8&14.3&15.7&13.4&17.3&\cellcolor[gray]{0.9}\textbf{2.9}&\cellcolor[gray]{0.9}\textbf{\textcolor{red}{2.4}}&\cellcolor[gray]{0.9}\textbf{\textcolor{red}{2.7}}&11.9&8.9&10.3&\cellcolor[gray]{0.9}\textbf{5.9}&\cellcolor[gray]{0.9}\textbf{4.3}&6.1&16.9&18.6&15.3\\
{Melanie(DDM(Online Boosting)) with source}&
12.0&15.4&13.9&12.7&6.3&9.4&17.4&18.3&18.4&\cellcolor[gray]{0.9}\textbf{2.9}&7.8&14.4&\cellcolor[gray]{0.9}\textbf{4.4}&5.1&\cellcolor[gray]{0.9}\textbf{\textcolor{red}{4.6}}&\cellcolor[gray]{0.9}\textbf{\textcolor{red}{5.6}}&7.1&8.8&18.1&22.7&23.7\\
{Melanie($HDDM_A$(Online Bagging)) with source}&
10.3&5.9&15.3&15.8&17.2&12.3&12.1&11.1&8.7&\cellcolor[gray]{0.9}\textbf{\textcolor{red}{2.6}}&3.6&4.1&7.8&10.8&10.8&\cellcolor[gray]{0.9}\textbf{7.0}&\cellcolor[gray]{0.9}\textbf{\textcolor{red}{3.9}}&5.0&19.6&17.6&12.9\\
{Melanie($HDDM_A$(Online Boosting)) with source}&
12.1&15.3&15.2&12.7&6.3&11.1&14.8&19.1&20.3&\cellcolor[gray]{0.9}\textbf{3.0}&7.7&15.5&\cellcolor[gray]{0.9}\textbf{4.7}&6.8&5.8&7.9&6.1&8.3&19.4&23.8&23.5\\
{Melanie(DDM(Online Bagging)) without source}&
15.5&9.1&15.3&18.3&14.8&15.3&12.5&12.1&11.4&16.3&11.1&17.2&18.6&14.4&15.5&14.1&14.9&13.3&13.0&13.6&14.4\\
{Melanie(DDM(Online Boosting)) without source}&
11.6&18.2&14.5&9.2&6.2&10.2&13.2&15.3&15.2&14.0&18.4&16.6&11.9&8.3&11.3&15.3&17.9&18.3&15.0&20.0&19.8\\
{Melanie($HDDM_A$(Online Bagging)) without source}&
17.8&10.1&7.0&16.2&18.2&14.2&11.3&10.8&7.6&18.6&12.7&10.4&18.0&17.7&14.6&13.3&13.2&9.0&13.6&12.8&10.9\\
{Melanie($HDDM_A$(Online Boosting)) without source}&
8.8&18.1&15.2&9.2&9.4&11.6&12.6&16.0&15.6&12.2&18.4&17.1&11.9&10.7&12.6&14.3&18.7&18.7&14.2&18.9&20.7\\
{DDM(Online Bagging)}&
19.0&12.6&10.1&18.6&17.1&18.7&12.6&14.2&10.8&19.7&15.3&12.6&19.0&17.1&18.5&14.6&16.6&12.6&15.8&19.5&18.7\\
{DDM(Online Boosting)}&
23.7&23.1&24.9&17.9&21.3&18.2&15.7&15.9&14.9&23.7&23.1&24.9&19.3&21.8&18.2&17.3&18.1&17.7&15.4&13.3&12.9\\
{$HDDM_A$(Online Bagging)}&
19.0&12.6&10.1&19.8&18.7&16.3&11.2&10.4&6.9&19.7&15.3&12.6&20.0&18.6&16.3&13.3&12.6&8.3&16.8&18.1&17.1\\
{$HDDM_A$(Online Boosting)}&
23.7&24.4&23.6&17.9&17.3&15.0&16.5&15.6&11.2&23.7&24.4&23.6&19.3&18.4&15.6&17.7&18.2&14.3&17.4&14.5&13.4\\
{Online Bagging}&
19.0&12.6&10.1&19.8&19.2&23.6&18.5&17.3&17.0&19.7&15.3&12.6&20.0&19.1&23.2&19.0&17.8&17.4&16.9&18.0&17.6\\
{Online Boosting}&
23.7&24.4&23.6&17.9&17.3&15.0&20.2&22.5&20.2&23.7&24.4&23.6&19.3&18.4&15.6&21.1&23.4&22.0&16.2&14.0&14.3\\
{Adaptive Random Forest(DDM)}&
16.9&21.1&19.0&17.7&16.3&20.9&11.5&13.9&17.0&17.9&21.2&20.0&19.0&17.1&20.9&13.7&16.6&19.3&12.1&4.9&6.2\\
{Adaptive Random Forest($HDDM_A$)}&
16.9&21.4&20.0&18.9&18.5&22.5&15.6&14.0&14.8&17.9&21.4&20.8&19.7&18.9&22.4&17.2&16.7&17.3&14.6&6.3&4.9\\
{Dynamic Weighted Majority}&
15.5&3.5&2.1&17.3&15.6&18.9&12.1&11.8&9.4&16.4&7.0&3.2&18.6&15.9&19.1&14.3&14.1&12.1&10.3&16.9&15.4\\
\hline
\end{tabular}}
\begin{tablenotes}
\item  Friedman's p-values were always $<2.2 \times 10^{-16}$. The best approach has its ranking in red with grey background and the approaches not significantly different from it according to the Nemenyi test are in bold with grey background. Mean accuracy and standard deviations are in the Supplementary Material \cite{supplementary}.
\end{tablenotes}
\label{tab:Rank}
\vspace{-0.45cm}
\end{table*}

\subsection{Comparison on Artificial Datasets}
\subsubsection{Experiments with Non-Similar Source}
The experiment aims to investigate whether or not the use of very different concept sources by MARLINE can help us to improve the predictive performance. The Friedman ranking of the approaches on each dataset is shown in Table \ref{tab:Rank}. We can see that MARLINE with source is amongst the best performers under different amounts of the training data and different types of drifts, as shown by the table cells highlighted in grey, except the incremental drifts with the class size of 500. MARLINE without source is sometimes amongst the best, demonstrating that mapping the new target concept to the space of the historical target concepts is also beneficial. Figure \ref{fig: NS} shows some representative results across time. Other figures were omitted due to space restrictions.

\subsubsection{Experiments with Similar Source}
Melanie was designed to transfer knowledge with similar sources and target concepts, being thus expected to achieve the best performance for these data streams. Based on Friedman and Nemenyi tests shown in Table \ref{tab:Rank}, Melanie outperforms the other approaches in most scenarios. However, MARLINE with source also achieves competitive results, similar to those of Melanie. In some cases, the performance of MARLINE with source is better than that of Melanie, e.g., MARLINE($HDDM_A$(Online Boosting)) with source and Melanie ($HDDM_A$(Online Boosting)) with source with the class size of 500 on the abrupt drift dataset.

\subsection{Comparison on Real World Datasets}

London and Washington D.C. bike sharing data were collected from different sources, so their input and output spaces are quite different (see Table \ref{tab:RealInputSpace}). Therefore, the concepts (both in terms of domains and tasks) of the two datasets are supposed to be very different. From Table \ref{tab:Rank}, MARLINE($HDDM_A$(Online Bagging)) with source has the best performance on all the real world datasets. MARLINE without source is the 2nd best. From Figure \ref{fig: Real}, we can see that the accuracy of MARLINE with source is quite similar to that of MARLINE without source. This may be due to the adaptive mechanism of MARLINE.

It is worth noting that when the concept of the stream was easier to learn (as for the artificial datasets), then MARLINE was most helpful in the beginning of the stream and right after drifts. This is because, with time increasing, every method can learn the concept well (Figure \ref{fig: NS}). However, when the concept was more complex (as in the real world datasets), then MARLINE provided great help throughout time (Figure~\ref{fig: Real}).
Additional related analyses are in the Supplementary Material~\cite{supplementary}.

\begin{table}[t]\centering
\caption{Input Space and Rental Count for Real World Dataset.}
\vspace{-0.2cm}
\scalebox{0.9}{
\begin{tabular}{|l|c|c|c|c|c|}
\hline
Feature
&{AT}
&{FT}
&{HD}
&{WS}
&{RC}\\
\hline
London&$-1.5:34$&$-6:34$&$20.5:100$&$0:56.5$&$0:7860$\\
Washington D.C&$0.02:1$&$0:1$&$0:1$&$0:0.85$&$1:977$\\
\hline
\end{tabular}}
\begin{tablenotes}
\item AT: Actual Temperature; FT Feeling Temperature; HD: Humidity; WS: Wind Speed; RC: Rental Count.
\end{tablenotes}
\label{tab:RealInputSpace}
\vspace{-0.45cm}
\end{table}

\subsection{Contribution of Source+ Sub-classifiers}
\begin{figure}[t]
\vspace{-0.2cm}
\begin{subfigure}{0.235\textwidth}
\centerline{\includegraphics[scale=0.2]{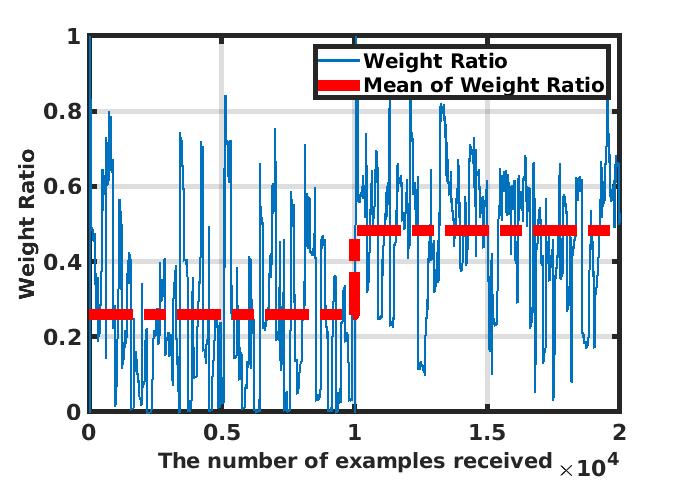}}
\caption{Non-similar; Abrupt; Size 5000}
\label{subfig:weight-e2s5000}
\end{subfigure}
\begin{subfigure}{0.235\textwidth}
\centerline{\includegraphics[scale=0.2]{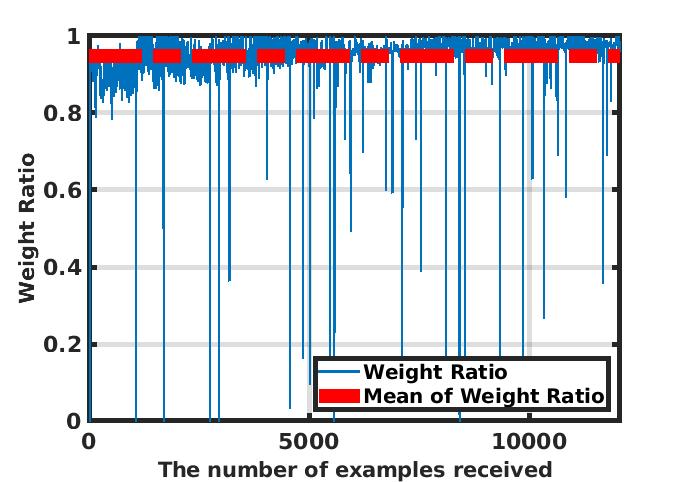}}
\caption{Weekday}
\label{subfig:weight-weekday}
\end{subfigure}
\caption{Sources+ Sub-classifiers' Average Total Weight (Over 30 Runs) by MARLINE ($HDDM_A$(Online Bagging)) with source.}
\label{fig: weight}
\vspace{-0.45cm}
\end{figure}

To further investigate the importance of individual sub-classifiers in MARLINE, we select two datasets (Abrupt with non-similar source and class size of 5000; and Weekday) and plot the average weight ratios of all the source+ sub-classifiers over 30 runs in Figure \ref{fig: weight}. The total weight ratio of the source+ sub-classifiers is calculated as:
\[    WeightRatio = \sum_{{i}\in\mathcal{M}, i \neq T}\sum_{{j}=1}^{J_{i}}\sum_{k=1}^{K}\omega_{h^{j,k}_{i}} + \sum_{{j}=1}^{J_{T}-1}\sum_{k=1}^{K}\omega_{h^{j,k}_{T}} \]
This is the sum of the total weights assigned to the source and historical target sub-classifiers in the MARLINE ensemble. 

From Figure \ref{subfig:weight-e2s5000}, before the concept drift occurs, the mean total weight of the source sub-classifiers during this period is 25.86\%. After concept drift, due to past target sub-classifiers joining the MARLINE ensemble, the importance of the source+ sub-classifiers increases and the mean total weight of the source+ sub-classifiers is 48.24\%.
The average total weight is even larger for the real world dataset (see Figure \ref{subfig:weight-weekday}). From Figure \ref{subfig:weight-weekday}, we notice that the source+ classifiers can significantly contribute towards the predictions throughout time (the mean of the total weights over the whole data stream is 94.88\%). This may be due to the fact that the real world dataset poses more challenges to the target sub-classifiers, which struggle to maintain their performance on the artificial datasets. We also notice some spikes and sudden drops in the total weights over time. This suggests that the weighting mechanism is affected by noise. In our future work, we will investigate whether or not other weighting mechanisms can improve the predictive performance of MARLINE further.

\section{Sensitivity Analysis}
As MARLINE has a few hyperparameters, it is important to conduct a study to understand (\textbf{Q1}) how different MARLINE ensemble compositions (different types of base learning ensemble with different sizes $K$ and performance index $\sigma$) affect the predictive performance, and (\textbf{Q2}) the influence of different drift detection methods and forgetting factors $\theta$ used to handle different types of concept drift.
Section \ref{subsec:sens_arti} answers these questions based on artificial datasets. The real world datasets are also used to support the analysis in Section \ref{subsec:sens_real}.

\subsection{Experimental Design}
\label{subsec: sens_design}
To investigate (\textbf{Q1}) and (\textbf{Q2}), Analysis of Variance (ANOVA) \cite{montgomery2017design} was performed to analyse the influence of each hyperparameter as well as its interactions with others on the average prequential accuracy. The step-wise changes of each hyperparameter are defined to cover the range of the best hyperparameter values selected by the grid search for different datasets in Section \ref{sec:Experiments}.

For (\textbf{Q1}), the following factors are investigated: base learner with two levels (BLM: Online Bagging and Boosting), base learning ensemble size $K \in \{10, 20, 30\}$ and performance index $\sigma \in \{0.0, 0.2, 0.4, 0.6\}$. As these are all subject-based factors, a Repeated Measures ANOVA design is used. For (\textbf{Q2}), the following factors are investigated:
drift detection method (DD: DDM and $HDDM_A$), forgetting factor $\theta \in \{0.9, 0.92, 0.94, 0.96, 0.98, 1\}$ and drift type (DT: No Drift, Abrupt, Incremental). The last is only considered when we apply the artificial datasets. As the first two factors are within-subject factors and the last is a between-subjects factor, a split plot (mixed) ANOVA design is adopted for the artificial datasets and a Repeated Measures ANOVA design is adopted for the real world datasets. Thirty runs for each combination of the factors are carried out on each dataset.

Mauchly's sphericity test \cite{mauchly1940significance} is used with a level of significance of 0.05 to evaluate whether or not the sphericity assumption is violated. If violated, the ANOVA's p-values are corrected to take that into account. If the epsilon estimate is below 0.75, the Greenhouse–Geisser correction \cite{greenhouse1959methods} is adopted to correct the degrees of freedom of the F-distribution. Otherwise, the Huynh–Feldt correction \cite{huynh1976estimation} is adopted to make it less conservative \cite{verma2015repeated}.

\subsection{Results}
Table \ref{tab:ANOVA_arti} and \ref{tab:ANOVA_real}
present the ANOVA results for the artificial and real world datasets, respectively.
The Sum of squares (SS), degrees of freedom (DF), mean squares (MS), test F statistics (F) and partial eta-squared ($\eta^2_p$) are reported.

\subsubsection{Analysis Using Artificial Datasets}
\label{subsec:sens_arti}

\begin{table}[t]
\caption{ANOVA Results for Artificial Datasets.}
\vspace{-0.2cm}
\begin{center}
\begin{tabular}{|l|l|l|l|l|l|}
\hline
Factor/Int.                       & SS     & DF    & MS     & F        & $\eta^2_p$   \\ \hline
\multicolumn{6}{|c|}{Test of Within-Subjects Effects (\textbf{Q1})}\\ \hline
$\sigma$                             & 26.142 & 1.012 & 25.826 & 2409.381 & 0.199 \\
$K$                     & 1.455  & 1.681 & 0.865  & 1487.022 & 0.133 \\
$K$ * $\sigma$             & 2.999  & 1.601 & 1.873  & 1468.858 & 0.131 \\
BLM * $K$         & 0.133  & 1.719 & 0.077  & 165.938  & 0.017 \\
BLM                         & 0.281  & 1     & 0.281  & 117.828  & 0.012 \\
BLM * $K$ * $\sigma$ & 0.099  & 1.702 & 0.058  & 57.919   & 0.006 \\
BLM * $\sigma$                 & 0.18   & 1.023 & 0.176  & 48.067   & 0.005 \\ \hline
\multicolumn{6}{|c|}{Test of Within-Subjects Effects (\textbf{Q2})}\\ \hline
DD * DT         & 0.9   & 2     & 0.45  & 796.4    & 0.076 \\
$\theta$                   & 7.64  & 1.191 & 6.413 & 1530.274 & 0.073 \\
$\theta$ * DT       & 7.427 & 2.382 & 3.117 & 743.804  & 0.071 \\
DD * $\theta$ * DT & 0.128 & 2.397 & 0.053 & 66.069   & 0.007 \\
DD                     & 0.023 & 1     & 0.023 & 40.876   & 0.002 \\
DD * $\theta$             & 0.01  & 1.198 & 0.008 & 10.508   & 0.001 \\ \hline
\multicolumn{6}{|c|}{Test of Between-Subjects Effects (\textbf{Q2})}\\ \hline
DT  & 553.298 & 2 & 276.649 & 12346.194 & 0.56 \\ \hline
\end{tabular}\end{center}
\vspace{-0.2cm}
\begin{tablenotes}
\item BLM: Base Learner Method; DD: Drift Detector Method; DT: Drift Type. The p-value is always less than 0.001, except for DD * $\theta$, which is 0.001.
\end{tablenotes}
\label{tab:ANOVA_arti}
\end{table}

\begin{figure}[t]
\begin{subfigure}{0.235\textwidth}
\centerline{\includegraphics[scale=0.15]{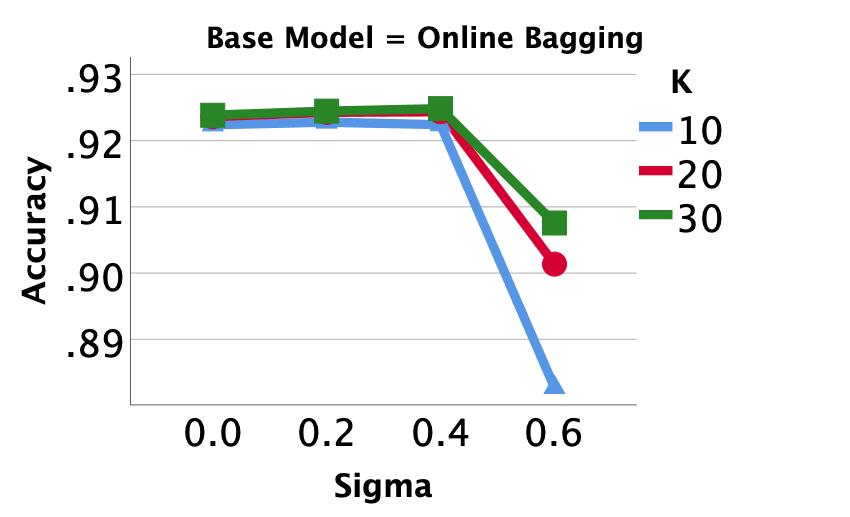}}
\caption{Effect of $K$*$\sigma$ with Online Bagging}
\label{subfig:sens_bag}
\end{subfigure}
\begin{subfigure}{0.235\textwidth}
\centerline{\includegraphics[scale=0.15]{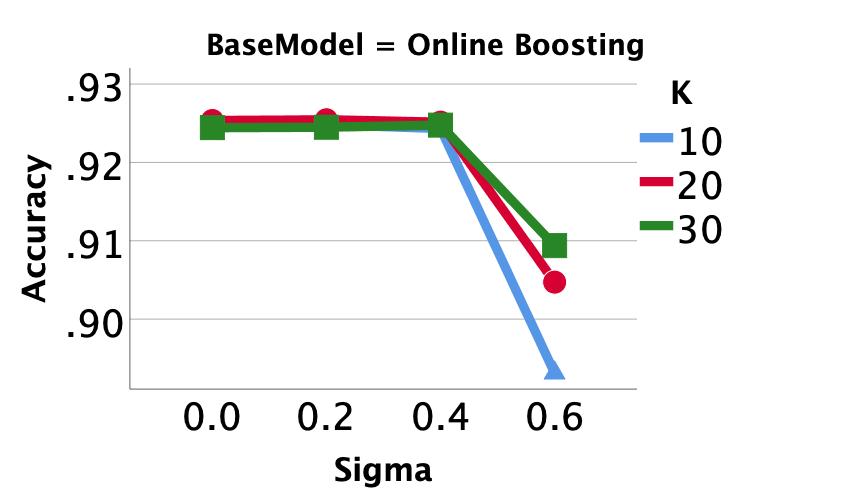}}
\caption{Effect of $K$*$\sigma$ with Online Boosting}
\label{subfig:sens_boost}
\end{subfigure}
\begin{subfigure}{0.235\textwidth}
\centerline{\includegraphics[scale=0.15]{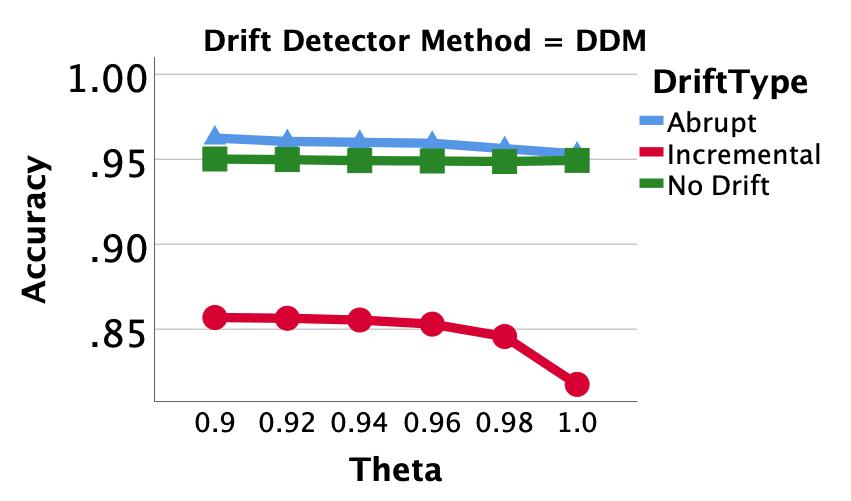}}
\caption{Effect of $\theta$*DT with DDM}
\label{subfig:sens_DDM}
\end{subfigure}
\begin{subfigure}{0.235\textwidth}
\centerline{\includegraphics[scale=0.15]{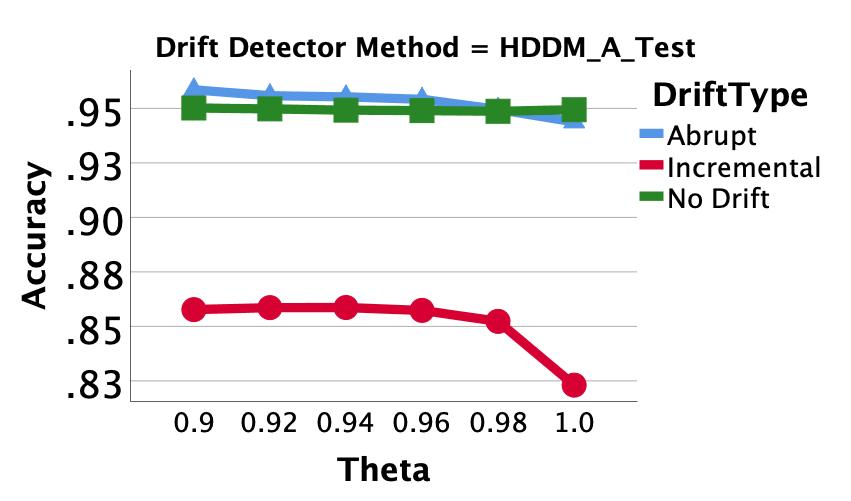}}
\caption{Effect of $\theta$*DT with $HDDM_A$}
\label{subfig:sens_hddm}
\end{subfigure}
\caption{Plots of marginal means on Artificial Datasets.}
\label{fig: sens_arti}
\vspace{-0.45cm}
\end{figure}

As it can be observed from Table \ref{tab:ANOVA_arti}, performance index $\sigma$, base learning ensemble size $K$ and interaction $K * \sigma$ have a large impact ($\eta^2_p \geq 0.131$), whereas the other factors and interactions have a small impact ($\eta^2_p \leq 0.017$). Therefore, the MARLINE ensemble composition factors $\sigma$ and $K$ have more influence on the accuracy.

Figures \ref{subfig:sens_bag} and \ref{subfig:sens_boost} illustrate the impact of factors $\sigma$, $K$, base ensemble learner method and their interaction. The two plots are fairly similar to each other, confirming that the interaction BLM * $K$ * $\sigma$ has a small impact.
A large $\sigma = 0.06$ is detrimental to the accuracy with worse accuracy obtained especially with a smaller ensemble size $K$. This is because this performance index is difficult to be reached by the sub-classifiers. Therefore, most sub-classifiers will have weight zero in the MARLINE ensemble, effectively decreasing its size and diversity. 
When $\sigma \leq 0.4$, different ensemble sizes $K$ and $\sigma$ values lead to similar accuracy, where a smaller ensemble size, e.g. $K=10$, leads to slightly worse accuracy and $\sigma = 0.4$ leads to slightly better accuracy when we use Online Bagging. 


Table \ref{tab:ANOVA_arti} also shows that the forgetting factor $\theta$, the interaction between the drift detecting method and the drift type DD * DT and the interaction $\theta$ * DT have a medium impact ($0.071 \leq \eta^2_p \leq 0.076$). So, the drift detection method and $\theta$ play important and probably diverse roles when handling different types of concept drifts. Figures \ref{subfig:sens_DDM} and \ref{subfig:sens_hddm} illustrate the effect of the factors $\theta$, DD and DT and their interaction. The two plots show fairly similar patterns, verifying that the interaction DD * $\theta$ * DT has a small impact.

When the drift appears abruptly, independent of the drift detection method (DT),  $\theta = 0.9$ result in the best accuracy. As $\theta$ increases, the accuracy slightly decreases. When the drift type is incremental,
the accuracy has larger drops with $\theta \geq 0.96$, compared with the abrupt concept drift. As the concept drifts in the incremental datasets are more difficult to be detected than the concept drifts in the abrupt datasets, it is reasonable that $\theta$ will take more responsibilities to cope with concept drifts when drift detection does not perform well. When the dataset has no drift, there is no significant difference between the accuracy obtained by different drift detection methods, which we confirmed by additional paired T tests with Bonferroni corrections. Furthermore, the accuracy changes very slightly when we change $\theta$. 


Therefore, we summarise that:
\begin{itemize}
\item \textbf{Q1:} 
Large performance index $\sigma$ and small ensemble sizes $K$ can be detrimental to the predictive performance of MARLINE, whereas in general $\sigma=0.4$ associated with $K \geq 20$ led to better results.
\item \textbf{Q2:} If there are concept drifts in the data steam, when the drift detection method cannot detect the concept drifts accurately, a small value for the forgetting factor ($0.9 \leq \theta \leq 0.94$) normally can help MARLINE to increase predictive performance on handling concept drifts. When there is no concept drift, a small value for the forgetting factor will not hurt the performance either.
\end{itemize}

\subsubsection{Analysis Using Real World Datasets}
\label{subsec:sens_real}

\begin{figure}[tb]
\begin{subfigure}{0.235\textwidth}
\centerline{\includegraphics[scale=0.15]{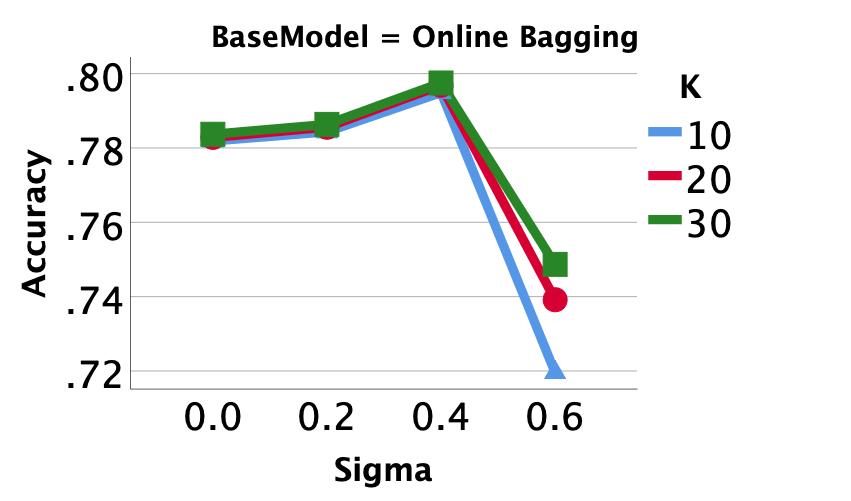}}
\caption{Effect of $K$*$\sigma$ with Online Bagging}
\label{subfig:sens_real_bag}
\end{subfigure}
\begin{subfigure}{0.235\textwidth}
\centerline{\includegraphics[scale=0.15]{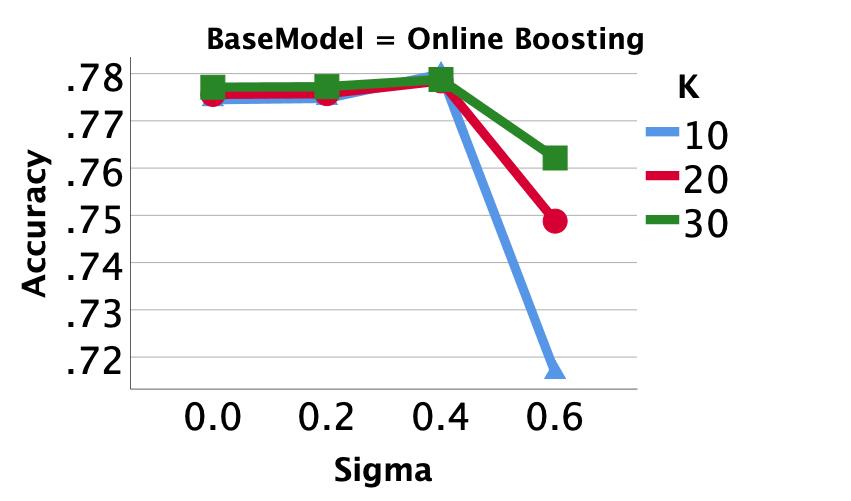}}
\caption{Effect of $K$*$\sigma$ with Online Boosting}
\label{subfig:sens_real_boost}
\end{subfigure}
\begin{subfigure}{0.235\textwidth}
\centerline{\includegraphics[scale=0.15]{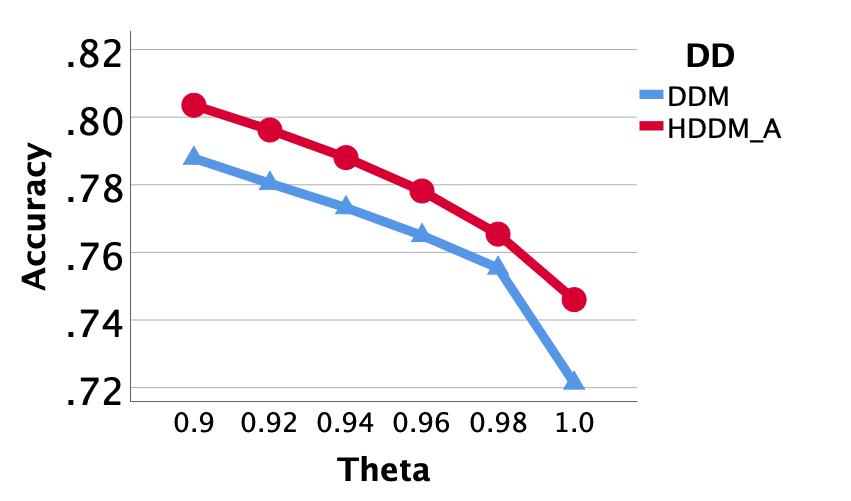}}
\caption{Effect of DD*$\theta$}
\label{subfig:sens_real_DDM}
\end{subfigure}
\caption{Plots of marginal means on Real World Datasets.}
\label{fig:sens_real}
\vspace{-0.45cm}
\end{figure}

Table \ref{tab:ANOVA_real} shows the tests of within-subjects performed on the real world datasets.
The plots of marginal means are shown in Figure \ref{fig:sens_real}. The results are in general similar to the results on artificial datasets, though certain effects and differences in the magnitude of the predictive performance were larger. We can see that $\sigma$, $K$ and interaction $\sigma * K$ have large effect size ($\eta^2_p \geq 0.161$). Meanwhile, $\theta$ has a very large effect size ($\eta^2_p = 0.495$) and the choice of drift detection method also has a large effect size ($\eta^2_p = 0.111$). This could be because in the real world datasets, the concept drifts are a mix of different types of concept drift, which makes them more difficult to be detected. Therefore, MARLINE relies more on $\theta$ to cope with the concept drifts.

\begin{table}[t]\centering
\caption{ANOVA Results for Real World Datasets.}
\vspace{-0.2cm}
\begin{center}
\begin{tabular}{|l|l|l|l|l|l|}
\hline
Factor/Int.                       & SS     & DF    & MS     & F        & $\eta^2_p$   \\ \hline
\multicolumn{6}{|c|}{Test of Within-Subjects Effects (\textbf{Q1})}\\ \hline
$\sigma$                             & 18.815 & 1.007 & 18.692 & 845.271  & 0.281 \\
$K$ * $\sigma$             & 2.16   & 1.695 & 1.275  & 461.576  & 0.176 \\
$K$                     & 1.008  & 1.699 & 0.593  & 413.163  & 0.161 \\
BLM                         & 0.6    & 1     & 0.6    & 231.762  & 0.097 \\
BLM * $\sigma$                 & 0.979  & 1.038 & 0.943  & 182.082  & 0.078 \\
BLM * $K$ * $\sigma$ & 0.14   & 1.909 & 0.074  & 36.82    & 0.017 \\
BLM * $K$         & 0.027  & 1.903 & 0.014  & 12.6     & 0.006 \\ \hline
\multicolumn{6}{|c|}{Test of Within-Subjects Effects (\textbf{Q2})}\\ \hline
$\theta$                             & 21.629 & 1.302 & 16.612 & 4235.103 & 0.495 \\
DD                               & 3.197  & 1     & 3.197  & 541.906  & 0.111 \\
DD * $\theta$                       & 0.26   & 1.194 & 0.218  & 98.49    & 0.022 \\ \hline
\end{tabular}
\end{center}
\vspace{-0.2cm}
\begin{tablenotes}
\item BLM: Base Learner Method; DD: Drift Detector Method; DT: Drift Type. The p-value is always less than 0.001.
\end{tablenotes}
\label{tab:ANOVA_real}
\end{table}

Figures \ref{subfig:sens_real_bag} and \ref{subfig:sens_real_boost} show similar  trends to the artificial datasets. However, when $\sigma \leq 0.4$, the improvement in accuracy with a greater $\sigma$ is more significant, confirming that both the size and the quality (performance index) of the sub-classifiers are important. Figure \ref{subfig:sens_real_DDM} also shows that $HDDM_A$ performs better on the real world datasets, in line with the experiments shown in Section \ref{sec:Experiments}. Also, we find that smaller $\theta$ values benefit the accuracy more.

\section{Conclusion}
In this paper, we focus on a general and challenging problem – learning from very different concepts in data stream mining. By mapping the target concept to the space of each source+ concept, the sub-classifiers that closely match the part of the projection of the target concept are given higher weights in the MARLINE ensemble, being able to achieve better performance in non-stationary environments. We carried out extensive experiments and the results demonstrate that our proposed MARLINE is effective. A sensitivity analysis is also presented. 
Future work includes the investigation of strategies to reduce the size of MARLINE's classifier pool; investigation of different weighting schemes to further improve accuracy; analysis of the computational time taken to run the approach, complementing its time complexity analysis; experiments with more data streams, base learners and drift detection methods; and an investigation of sensitivity to noise.

\bibliographystyle{IEEEtran}
\bibliography{ref}

\end{document}